\documentclass[lettersize,journal]{IEEEtran}
\usepackage{amsmath,amsfonts}
\usepackage{algorithm}
\usepackage{algpseudocode}
\usepackage{array}
\usepackage[caption=false,font=normalsize,labelfont=sf,textfont=sf]{subfig}
\usepackage{textcomp}
\usepackage{stfloats}
\usepackage{url}
\usepackage{verbatim}
\usepackage{graphicx}
\usepackage{booktabs}
\usepackage{multirow}
\usepackage{cite}
\usepackage{orcidlink}

\begin{document}

\title{MDM: Advancing Multi-Domain Distribution Matching for Automatic Modulation Recognition Dataset Synthesis}

\author{Dongwei Xu\orcidlink{0000-0003-2693-922X},~\IEEEmembership{Member, IEEE}, Jiajun Chen, Yao Lu\orcidlink{0000-0003-0655-7814}, Tianhao Xia, Qi Xuan\orcidlink{0000-0002-6320-7012},~\IEEEmembership{Senior Member,~IEEE}, Wei Wang, Yun Lin, Xiaoniu Yang\orcidlink{0000-0003-3117-2211}
\thanks{This work was supported by the National Natural Science Foundation of China under Grants U21B2001. \textit{(Corresponding author: Dongwei Xu.)}}
\thanks{Dongwei Xu, Jiajun Chen, Yao Lu, Tianhao Xia and Qi Xuan are with the Institute of Cyberspace Security, College of Information Engineering, Zhejiang University of Technology, Hangzhou 310023, China (e-mail: dongweixu@zjut.edu.cn; 221122030156@zjut.edu.cn; yaolu.zjut@gmail.com;
1026826124@qq.com; xuanqi@zjut.edu.cn).}
\thanks{Wei Wang is with the Institute of China Electronics Technology Group Corporation 36th Research Institute, Jiaxing 314033. China (e-mail: wwwzwh@163.com).}
\thanks{Yun Lin is with the Institute of College of Information And Communication Engineering, Harbin Engineering University, Harbin 150001. China (e-mail: linyun@hrbeu.edu.cn).}
\thanks{Xiaoniu Yang is with the Institute of Cyberspace Security, Zhejiang University of Technology, Hangzhou 310023, China, and also with the National Key Laboratory of Electromagnetic Space Security, Jiaxing 314033. China (e-mail: yxn2117@126.com).}}

\markboth{Journal of \LaTeX\ Class Files,~Vol.~14, No.~8, August~2021}%
{Shell \MakeLowercase{\textit{et al.}}: A Sample Article Using IEEEtran.cls for IEEE Journals}


\maketitle

\begin{abstract}
Recently, deep learning technology has been successfully introduced into Automatic Modulation Recognition (AMR) tasks. However, the success of deep learning is all attributed to the training on large-scale datasets. Such a large amount of data brings huge pressure on storage, transmission and model training. In order to solve the problem of large amount of data, some researchers put forward the method of data distillation, which aims to compress large training data into smaller synthetic datasets to maintain its performance. While numerous data distillation techniques have been developed within the realm of image processing, the unique characteristics of signals set them apart. Signals exhibit distinct features across various domains, necessitating specialized approaches for their analysis and processing. To this end, a novel dataset distillation method---Multi-domain Distribution Matching (MDM) is proposed. MDM employs the Discrete Fourier Transform (DFT) to translate time-domain signals into the frequency domain, and then uses a model to compute distribution matching losses between the synthetic and real datasets, considering both the time and frequency domains. Ultimately, these two losses are integrated to update the synthetic dataset. We conduct extensive experiments on three AMR datasets. Experimental results show that, compared with baseline methods, our method achieves better performance under the same compression ratio. Furthermore, we conduct cross-architecture generalization experiments on several models, and the experimental results show that our synthetic datasets can generalize well on other unseen models.
\end{abstract}


\section{Introduction}
\IEEEPARstart{I}{n} recent years, deep learning techniques have been gradually introduced into AMR tasks~\cite{chen2021signet,hou2023signal,shangao2022learning,zhang2022amr,ryu2023emc} and have achieved great success. The success of deep learning is due to training on large datasets, but such a large amount of data poses a huge challenge for storage and transmission, and brings a significant amount of time and computing resource overhead for model training. In order to solve this problem, some researchers have proposed a promising direction named dataset distillation (DD), which aims to use limited data to train the model in a more efficient way, thereby reducing the training cost and improving the model performance. 

DD was firstly proposed by Wang et al.~\cite{Dataset} in 2018, which viewed model parameters as a function about the synthetic dataset and updated the synthetic dataset by minimizing the training loss of the original training data with respect to the synthetic data. Subsequently, Zhao et al.~\cite{zhao2020dataset} proposed a method named Dataset Condensation (DC). This method updated the synthetic dataset by matching the gradient between the real training set and the synthetic dataset. 
Zhao et al.~\cite{zhao2021dataset} proposed Differentiable Siamese Augmentation (DSA), which brought data augmentation technique to DC and achieved better performance.
Zhao et al.~\cite{zhao2023dataset} proposed a method based on Distributed Matching (DM), which used a model to extract high-dimensional features of the synthetic dataset and the original training set, and updated the synthetic dataset by matching the distance between the two high-dimensional features. 
Lu et al.~\cite{lu2023can} introduced two plug-and-play loss terms, CLoM and CCloM, which provide stable guidance for optimizing synthetic datasets. 
Although many data distillation techniques have been developed for image processing, signals have unique attributes like temporal dynamics and frequency characteristics that set them apart. These differences mean that conventional image processing techniques may not be effective for signals, requiring specialized methods tailored to the specific nature of signals for AMR tasks.

To this end, this paper combines the time domain and frequency domain of the signal to achieve the complementary gain of the information in the two domains. 
The time domain signal is first mapped to the frequency domain using the DFT~\cite{jenkins2022fourier}, and the DM loss between the real training set and the synthetic dataset is then computed in both time and frequency domains. The two losses are combined as the final objective function. By minimizing the objective function, the optimal synthetic dataset is achieved. Experiments are conducted on three modulated signal datasets: RML2016.10a-high, RML2016.10a~\cite{o2016radio} and Sig2019-12-high~\cite{chen2021signet}. Additionally, cross-architecture generalization experiments are carried out on the AlexNet~\cite{krizhevsky2012imagenet} and VGG16~\cite{simonyan2014very} models. Random selection, Forgetting~\cite{toneva2018empirical}, DC~\cite{zhao2020dataset} and DM~\cite{zhao2023dataset} are taken as baseline methods. Experiments show that compared with baseline methods, MDM performs better under the simliar conditions. At the same time, the synthetic dataset learned by MDM has certain generalization performance. The main contributions of this work are as follows: 
\begin{itemize}
  \item[$\bullet$] An innovative dataset distillation technique called MDM is introduced. Specifically, MDM transforms the time domain signal into the frequency domain using DFT and integrates the characteristics of both time and frequency domains to distill the data.
  \item[$\bullet$] The performance of MDM is evaluated on three signal datasets, and it is compared with Random Select, Forgetting, DC, and DM methods. Experiments demonstrate that MDM achieves optimal classification accuracy in most cases.  
  \item[$\bullet$] Cross-architecture generalization experiments are conducted on multiple models, and the experimental results indicate that the synthetic dataset learned by MDM can generalize well on previously unseen model architectures.
\end{itemize}

\begin{figure*}
\centering
\includegraphics[width=\linewidth]{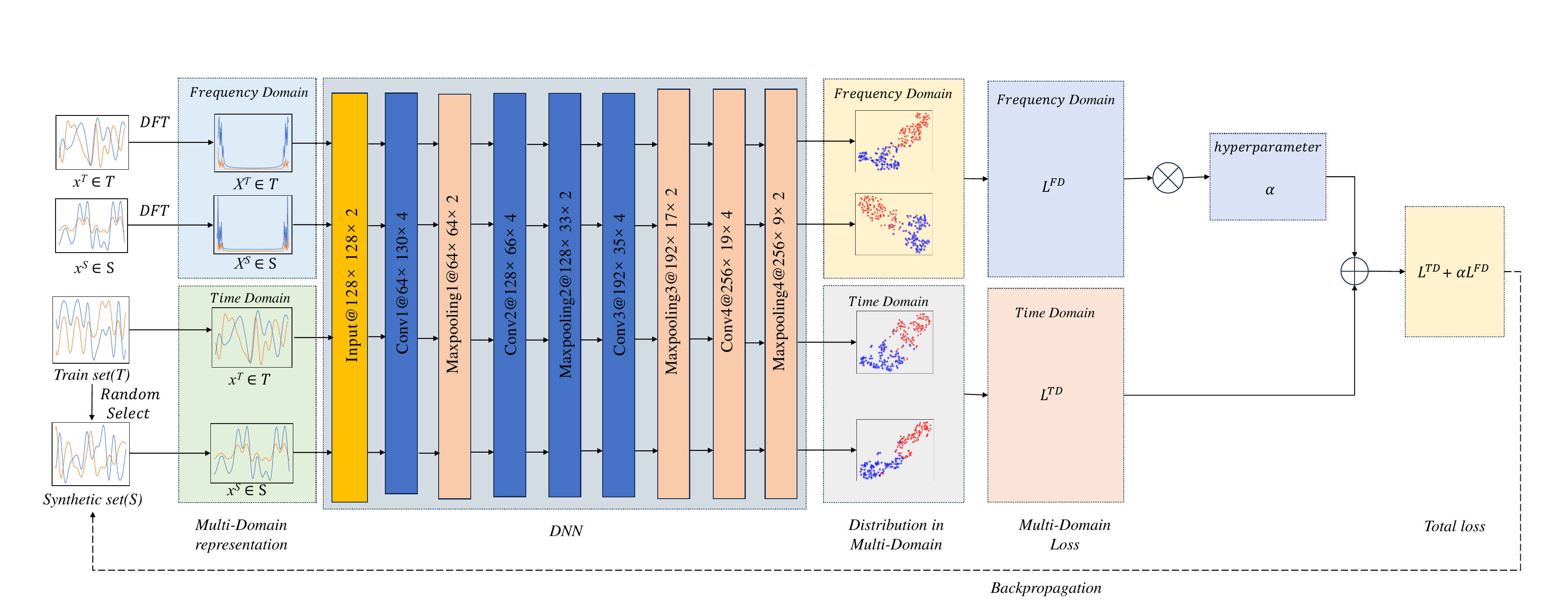}

\caption{\label{framework}The framework of Multi-domain Distribution Matching.}
\end{figure*}

\section{Multi-domain Distribution Matching}
\subsection{Dataset Condensation Problem}
A large-scale training dataset and a small synthetic dataset can be expressed as follow:
\begin{equation}
\begin{aligned}
    T &= \{(x_1^T, y_1^T), . . . ,(x_{|T|}^T, y_{|T|}^T)\}
\end{aligned}
\end{equation}
\begin{equation}
\begin{aligned}
    x^T=[I^T(n),Q^T(n)], n=0,1,...,N-1
\end{aligned}
\end{equation}
\begin{equation}
\begin{aligned}
    S &= \{({x}^{S}_1,{y}^{S}_1), . . . ,({x}^{S}_{|S|}, {y}^{S}_{|S|})\}
\end{aligned}
\end{equation}
\begin{equation}
\begin{aligned}
    {x}^{S}=[I^S(n),Q^S(n)], n=0,1,...,N-1
\end{aligned}
\end{equation}
where $|T|$ indicates that the training dataset $T$ contains $|T|$ signals and labels; $x^T$ represents the signal in the training dataset; each signal $x^T$ is represented by an I channel and a Q channel; each channel has $N$ sampling points; $|S|$ indicates that the synthetic dataset $S$ contains $|S|$ signals and labels; ${x^{S}}$ denotes the signal of the synthetic dataset, and $|T|\gg|S|$.

\begin{table*}[hb]
  \centering
  \caption{Testing Accuracy($\%$) comparing to coreset selection and training set synthesis methods. }
    \begin{tabular}{cccccccccc}
    \toprule
    & ~ & RML2016.10a-high  & ~ & ~ & RML2016.10a & ~ & ~ & Sig2019-12-high & ~ \\
    \midrule
    IPC & 10 & 50 & 100 & 10 & 50 & 100 & 10 & 50 & 100 \\ 
    Ratio(\%) & 0.25 & 1.25 & 2.5 & 0.0625 & 0.3125 & 0.625 & 0.1 & 0.5 & 1 \\ 
    \midrule
    Random & 55.8±1.0 & 71.8±1.2 & 75.8±0.8 & 25.3±2.0 & 33.8±2.1 & 38.9±2.2 & 25.6±1.0 & 42.5±0.7 & 53.5±1.4 \\ 
    Forgetting & 55.5±0.8 & 75.0±0.6 & \textbf{78.6±0.4} & 22.9±0.7 & 33.0±0.6 & 35.8±1.3 & 26.7±1.6 & 43.0±0.5 & 54.9±1.7 \\ 
    DC & 57.5±1.2 & 71.4±0.2 & 74.1±0.2 & 25.3±0.5 & 33.9±0.7 & 37.2±0.9 & 25.7±1.0 & 37.7±2.2 & 50.8±1.1 \\ 
    DM    & 61.3±0.8 & 74.8±0.2 & 76.6±1.0 & 28.9±0.7 & 38.6±0.6 & 42.8±0.4 & 29.1±1.2 & 47.5±0.6 & 56.2±0.7 \\
    MDM & \textbf{62.5±1.4} & \textbf{76.0±0.4} & 78.1±0.2 & \textbf{30.1±0.6} & \textbf{41.9±0.8} & \textbf{43.8±0.7} & \textbf{31.0±1.3} & \textbf{48.4±1.1} & \textbf{57.0±0.8} \\
     \midrule
    Whole dataset & \multicolumn{3}{c}{88.02±0.2} & \multicolumn{3}{c}{56.4±0.2} & \multicolumn{3}{c}{96.5±0.2} \\
    \bottomrule
    \end{tabular}%
  \label{tab:addlabel}%
\end{table*}%

\subsection{Distribution Matching in Time Domain}
First, a neural network model $\psi_\theta$ is randomly initialized, which is used to obtain low-dimensional embeddings of real signal $x^T$ and synthetic signal ${x^{S}}$. Then the maximum mean discrepancy (MMD)~\cite{gretton2012kernel} is used to estimate the distance between the real data distribution and the synthetic data distribution. Finally, this distance is taken as a loss function in the time domain, which can be expressed as:
\begin{equation}
\begin{aligned}
    L^{TD}=E_{\theta\sim{P_\theta}}\parallel{\frac{1}{\mid{T}\mid}\sum_{j=1}^{\mid{T}\mid}\psi_{\theta}(x_j^T)-\frac{1}{\mid{S}\mid}\sum_{m=1}^{\mid{S}\mid}\psi_{\theta}({x^{S}_m)}\parallel}^2
\end{aligned}
\end{equation}
where $P_{\theta}$ represents the distribution of network parameters; TD indicates Time Domain.

\begin{algorithm}[t]
    \caption{Multi-domain Distribution Matching}
    \label{algorithm:Partition}
    \textbf{Input}: Training set $T$, randomly initialized set of synthetic samples $S$, deep neural network $\psi_\theta$ parameterized with $\theta$, probability distribution over parameters $P_\theta$, training iterations $K$, learning rate $\eta$, tunable hyperparameter $\alpha$.\\
    \textbf{Output}: $S$.\\
    \begin{algorithmic}[1] 
        \For{$k = 0$ to $K-1$}
            \State Sample $\theta\sim{P_\theta}$
            \State Compute $L^{TD}=E_{\theta\sim{P_\theta}}\parallel\frac{1}{\mid{T}\mid}\sum_{j=1}^{\mid{T}\mid
            }\psi_{\theta}(x_j^T)-\frac{1}{\mid{S}\mid}\sum_{m=1}^{\mid{S}\mid
            }\psi_{\theta}({x}^S_m)\parallel^2$
            \State Perform DFT on $x_j^T$ and ${x}^{S}_m$:
            \State    ${X}^T_j=DFT(x_j^T)$,
                ${X}^{S}_m=DFT({x}^{S}_m)$
            \State Compute $L^{FD}=E_{\theta\sim{P_\theta}}\parallel\frac{1}{\mid{T}\mid}\sum_{j=1}^{\mid{T}\mid
            }\psi_{\theta}({X}^T_j)-\frac{1}{\mid{S}\mid}\sum_{m=1}^{\mid{S}\mid
            }\psi_{\theta}({X}^S_m)\parallel^2$
            \State Combine two losses $L=L^{TD}+\alpha L^{FD}$
            \State Update $S\leftarrow S-
            \eta{\nabla_S{L}}$
        \EndFor
    \end{algorithmic}
\end{algorithm}

\subsection{Discrete Fourier Transform}
DFT is a basic method in signal analysis that transforms a discrete time series signal from the time domain to the frequency domain.

In order to align with the time domain signal, DFT is performed on $I^T(n)$ and $Q^T(n)$ of signal $x^T$ in the training dataset, and they are then spliced together. The frequency domain signals of the two channels are also obtained, which can be expressed as follows:
\begin{equation}
\begin{aligned}
    I^T(k)=|\sum_{n=0}^{N-1}I^T(n)e^{-j2{\pi}kn/N}|,k=0,1,...,N-1,
\end{aligned}
\end{equation}
\begin{equation}
\begin{aligned}
    Q^T(k)&=|\sum_{n=0}^{N-1}Q^T(n)e^{-j2{\pi}kn/N}|,
\end{aligned}
\end{equation}
\begin{equation}
\begin{aligned}
    X^T=[I^T(k), Q^T(k)],
\end{aligned}
\end{equation}
where $X^{T}$ represents the frequency domain representation of each signal in the training dataset; each signal $X^{T}$ has two channels: $I^{T}(k)$ and $Q^{T}(k)$; each channel has $N$ sampling points (the input of the DFT is $N$ discrete points and the output of the DFT is $N$ discrete points). The same operation is performed on the synthetic dataset, which can be expressed as:
\begin{equation}
\begin{aligned}
    X^{S}=[I^{S}(k), Q^{S}(k)]
\end{aligned}
\end{equation}
where $X^{S}$ denotes the frequency domain representation of each signal in the synthetic dataset; $I^{S}(k)$ is the output of $I^{S}(n)$ after DFT, and $Q^{S}(k)$ is the output of $Q^{S}(n)$ after DFT.

\subsection{Distribution Matching in Fequency Domain}
Then, the low-dimensional embeddings of real signal $X^T$ and synthetic signal $X^S$ from the frequency domain are obtained by using $\psi_\theta$. MMD is used to estimate the distance between the real data distribution and the synthetic data distribution in the frequency domain. The distance is used as a loss function in the frequency domain, which can be expressed as:
\begin{equation}
\begin{aligned}
    L^{FD}=E_{\theta\sim{P_\theta}}\parallel{\frac{1}{\mid{T}\mid}\sum_{j=1}^{\mid{T}\mid}\psi_{\theta}(X_j^T)-\frac{1}{\mid{S}\mid}\sum_{m=1}^{\mid{S}\mid}\psi_{\theta}({X^{S}_m)}\parallel}^2&
\end{aligned}
\end{equation}
where FD means Frequency Domain. The framework of MDM is shown in the Fig. \ref{framework}.

\subsection{Combination of Time Domain and Frequency Domain}

Inspired by~\cite{qi2020automatic}, we believe that combining the time and frequency domain information of the signal is better than using only a single signal domain information.

Therefore, we combine the distribution matching loss in the time domain and frequency domain as the total loss:
\begin{equation}
\label{eq12}
\begin{aligned}
    L=L^{TD}+\alpha L^{FD}
\end{aligned}
\end{equation}
where $\alpha$ is a tunable hyeprparameter. The value of $\alpha$ in the experiment depends on the dataset. The overall pipeline is summarized in Algorithm \ref{algorithm:Partition}.

\begin{figure*}[hb]
\centering
\includegraphics[width=\linewidth]{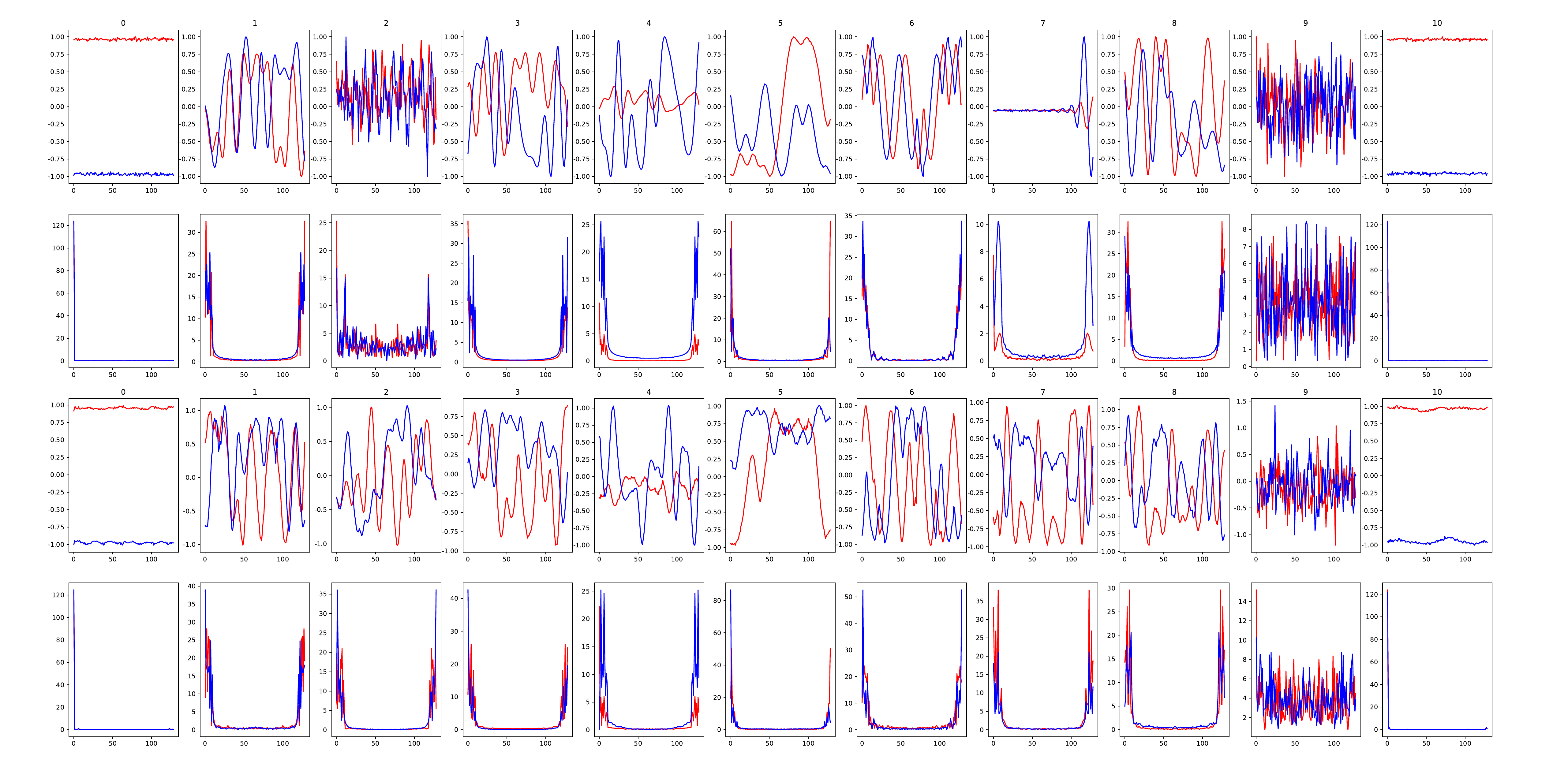}
\caption{Visualization of condensed 50 signals/class using AlexNet on RML2016.10a-high.}
\label{dataset:128_high}
\end{figure*}

\section{Experiments}
\subsection{Dataset}
To evaluate the effectiveness of our method, we conducted experiments on three signal modulation classification datasets RML2016.10a, RML2016.10a-high, and Sig2019-12-high.

\textbf{RML2016.10a} uses GNU radio to synthesize electromagnetic signals containing I and Q channels. There are a total of 11 modulation signal categories in the dataset, QPSK, 8PSK, BPSK, BFSK, 16QAM, 64QAM, CPFSK, PAM4, AM-SSB, WB-FM and AM-DSB, the first 8 types are digital modulation types, and the last 3 types are analog modulation types. The dataset uses an electromagnetic signal with a signal-to-noise ratio (SNR) range from -20db to 18dB and a signal length of 128. 
We divide the dataset into a training set and a test set with a ratio of 4:1.

\textbf{RML2016.10a-high} is the part of RML2016.10a dataset with SNRs ranging from 10 dB to 18 dB.
We also divide the dataset into the training set and the test set with a ratio of 4:1.

\textbf{Sig2019-12-high} is a subset of Sig2019-12, which is a self-generated dataset by Chen et al.~\cite{chen2021signet}. Sig2019-12 contains 12 modulation types, namely OPSK, 8PSK, BPSK, 4PAM, 8PAM, OQPSK, 16QAM, 32QAM, 64QAM, 2FSK, 4FSK and 8FSK. The signal-to-noise ratio of the dataset ranges from -20db to 30db.
~\cite{lu2024generic}, we select signals above 10dB, comprising a training set of 120,000 signal samples and a test set with 60,000 signal samples.

\subsection{Model Architecture}
In our experiments, an AlexNet model is used to learn synthetic dataset. For the evaluation phase, $5$ AlexNet models are randomly initialized and trained from scratch using the synthesized dataset that has been generated. Their average test accuracy and standard deviation are then recorded. 


\begin{table}[t]
    \centering
    \caption{Cross-architecture performance ($\%$) with condensed $50$ signal/class on RML2016.10a-high.}
    \begin{tabular}{cccccc}
    \toprule
        C$\backslash$T & AlexNet & 2D-CNN & VGG16 & 1D-ResNet & MCLDNN \\ 
        \midrule
        AlexNet & 74.7±1.0 & 59.7±0.6 & 71.2±2.2 & 52.2±2.2 & 51.7±3.5 \\
        VGG16 & 73.3±1.2 & 57.9±1.0 & 70.2±1.2 & 50.2±6.6 & 46.6±2.6 \\
        \bottomrule
    \end{tabular}
    \label{tab:table2}%
\end{table}

Conduct cross-architecture generalization experiments are further conducted. A synthetic dataset is first trained on one model architecture, and this dataset is then employed to train models across various other architectures. The architectures used in the cross-architecture generalization experiments include AlexNet~\cite{krizhevsky2012imagenet}, 1D-ResNet~\cite{o2018over}, 2D-CNN~\cite{o2016convolutional}, VGG16~\cite{simonyan2014very}, and MCLDNN~\cite{xu2020spatiotemporal}. This allows us to evaluate the generalization performance of the synthetic datasets across different model architectures.
 
\subsection{Implementation details}
At the beginning, some training samples from each class in the training set are randomly selected as the initial synthetic samples. Then, the synthetic dataset is updated. Specifically, for RML2016.10a-high dataset, the optimal learning rate is $10^{-5}$ when signal per class (SPC) =$10$, and the optimal learning rate is $10^{-4}$ when SPC=$50$ and $100$. For RML2016.10a dataset, the optimal learning rate is $10^{-8}$ when SPC=$10$, when SPC=$50$ and $100$, the optimal learning rate is $10^{-7}$. For Sig2019-12-high dataset, the optimal learning rate is $10^{-6}$ when SPC=$10$, when SPC=$50$, the optimal learning rate is $10^{-5}$, and when SPC=$100$, the optimal learning rate is $10^{-4}$. The total epochs of iterations is $20,000$. Next, a randomly initialized model is trained on the updated synthetic dataset. The learning rate of the model training is $0.001$, the optimizer is SGD, the momentum is $0.9$, the sample batch size is $128$, and the epochs of training iterations are $300$.


\subsection{Results and Analysis}
Our method is compared with several baseline methods: Random Selection, Forgetting, DC, and DM. These baselines are briefly summarized as follows:
\begin{itemize}
  \item[$\bullet$] Random Selection: This algorithm randomly selects a certain number of training samples from each class of training samples as synthetic samples.
  \item[$\bullet$] Forgetting: This method counts how many times a training sample is learned and then forgotten during network training. The samples that are less forgetful can be dropped. 
  \item[$\bullet$] DC: This method synthesizes a small but informative dataset by matching the loss gradients with respect to the training and synthetic datasets.
  \item[$\bullet$] DM: This method aims to minimize the MMD between the synthetic dataset and the real dataset.
\end{itemize}


\textbf{Comparison to coreset selection methods.}
Our method is first compared with the coreset selection baselines. As shown in Table \ref{tab:addlabel}, only when the dataset is RML2016.10a-high, SPC=100, Forgetting method has better results than MDM. In other cases, MDM is superior to Random Selection method and Forgetting method. This indicates that MDM generally surpasses coreset selection methods in most cases.

\textbf{Comparison to DC and DM.} 
As can be seen from Table \ref{tab:addlabel}, MDM is superior to the DC and DM in all cases. In most cases, our method is more than $1\%$ better than DM. These results illustrate that MDM, by incorporating frequency domain insights, capitalizes on the synergistic relationship between the time and frequency domains. This approach yields superior outcomes compared to DM and DC, which relies solely on time-domain signal information.

\subsection{Cross-Architecture Generalization.}
Cross-architecture generalization experiments are also conducted. Specifically, for the RML2016.10a-high dataset, 50 condensed signals per class are synthesized using two distinct models: AlexNet and VGG16. These synthesized signals are then evaluated across multiple architectures, including AlexNet, 1D-ResNet, 2D-CNN, VGG16, and MCLDNN. In Table \ref{tab:table2}, the synthetic dataset is learned with one architecture (C) and then evaluated on another architecture (T) by training a model from scratch and testing on real testing data. The experimental results show that the synthetic dataset trained on AlexNet and VGG16 performs best on AlexNet, with strong performance on VGG16 as well. It also achieves promising results on 1D-ResNet, 2D-CNN, and MCLDNN, indicating strong cross-model generalization.

\subsection{Visualization}
\label{sec: Visualization}
The training set and synthetic dataset are visualized in Fig. \ref{dataset:128_high}, showing time and frequency domain diagrams for each signal type when SPC=50 on the RML2016.10a-high dataset. The top rows display the training set diagrams, while the bottom rows show the synthetic dataset. The characteristics of the synthetic samples closely resemble those of the training set.

\section{Conclusion}
\label{sec:Conclusion}
In this paper, we propose a novel dataset distillation method named MDM. MDM maps the time domain signal to the frequency domain by DFT, and combines the time domain and frequency domain characteristics of the signal to distill the data, which makes up the gap of the dataset distillation method in the task of signal modulation recognition. Experiments reveal that the proposed method outperforms existing baselines and can generalize well on other unseen models.

\bibliographystyle{IEEEtran}
\bibliography{reference}











\newpage

\vfill

\end{document}